%% file: main.tex
\documentclass[dvipsnames,format=sigconf,anonymous=false,review=false,nonacm]{acmart}

\AtBeginDocument{
  }

\usepackage{subcaption}
\usepackage{booktabs}
\usepackage{tabularx}
\usepackage{cleveref}
\usepackage{siunitx}
\usepackage{tikz}
\usepackage{amsmath}
\usepackage[inline]{enumitem}

\newcolumntype{Y}{>{\centering\arraybackslash}X}
\usetikzlibrary{decorations.pathreplacing, positioning, calc, shapes.geometric}

\acmConference[GECCO '26]{The Genetic and Evolutionary Computation Conference 2026}{July 13--17, 2026}{San José, Costa Rica}
\acmYear{2026}
\copyrightyear{2026}

\begin{document}

\title{Evolutionary Mapping of Neural Networks to Spatial Accelerators}

\author{Alessandro Pierro}
\orcid{0000-0002-5682-627X}
\affiliation{%
  \institution{LMU Munich, Intel}
  \city{Munich}
  \country{Germany}
}
\email{alessandro.pierro@ifi.lmu.de}

\author{Jonathan Timcheck}
\orcid{0000-0002-2071-2668}
\affiliation{%
  \institution{Intel Corporation}
  \city{Santa Clara}
  \state{California}
  \country{USA}
}

\author{Jason Yik}
\orcid{0009-0009-5860-0619}
\affiliation{%
  \institution{SEAS, Harvard University}
  \city{Cambridge}
  \state{Massachusetts}
  \country{USA}}

\author{Marius Lindauer}
\orcid{0000-0002-9675-3175}
\affiliation{%
  \institution{L3S Research Center, Leibniz University Hannover}
  \city{Hannover}
  \country{Germany}
}

\author{Eyke H{\"u}llermeier}
\orcid{0000-0002-9944-4108}
\affiliation{%
  \institution{MCML, LMU Munich, DFKI}
  \city{Munich}
  \country{Germany}
}

\author{Marcel Wever}
\orcid{0000-0001-9782-6818}
\affiliation{%
 \institution{L3S Research Center, Leibniz University Hannover}
 \city{Hannover}
 \country{Germany}}

\renewcommand{\shortauthors}{Pierro et al.}

\begin{abstract}
Spatial accelerators, composed of arrays of compute-memory integrated units, offer an attractive platform for deploying inference workloads with low latency and low energy consumption.
However, fully exploiting their architectural advantages typically requires careful, expert-driven mapping of computational graphs to distributed processing elements.
In this work, we automate this process by framing the mapping challenge as a black-box optimization problem.
We introduce the first evolutionary, hardware-in-the-loop mapping framework for neuromorphic accelerators, enabling users without deep hardware knowledge to deploy workloads more efficiently.
We evaluate our approach on Intel Loihi 2, a representative spatial accelerator featuring 152 cores per chip in a 2D mesh. 
Our method achieves up to $35\%$ reduction in total latency compared to default heuristics on two sparse multi-layer perceptron networks. 
Furthermore, we demonstrate the scalability of our approach to multi-chip systems and observe an up to $40\%$ improvement in energy efficiency, without explicitly optimizing for it.
\end{abstract}

\ccsdesc[500]{Hardware~Power and energy}
\ccsdesc[500]{Computer systems organization~Parallel architectures}

\keywords{Evolutionary Algorithms, Resource Mapping, Partitioning and Placement, Hardware-aware Optimization, Combinatorial Optimization}


\maketitle

\input{Chapters/1_Introduction}

\input{Chapters/2_Background}

\input{Chapters/3_ProblemDefinition}

\input{Chapters/4_Algorithm}
\input{Chapters/5_Results}

\input{Chapters/6_Conclusions}

\begin{acks}
Marcel Wever and Marius Lindauer gratefully acknowledge funding by the European Union (ERC, ``ixAutoML'', grant no. 101041029). Views and opinions expressed are, however, those of the authors only and do not necessarily reflect those of the European Union or the European Research Council Executive Agency. Neither the European Union nor the granting authority can be held responsible for them.

\end{acks}

\bibliographystyle{ACM-Reference-Format}
\bibliography{strings,lib,proc,ext_lib}

\end{document}

%% file: Chapters/1_Introduction.tex
\section{Introduction}

\begin{figure}
    \centering
    \includegraphics[width=0.90\linewidth]{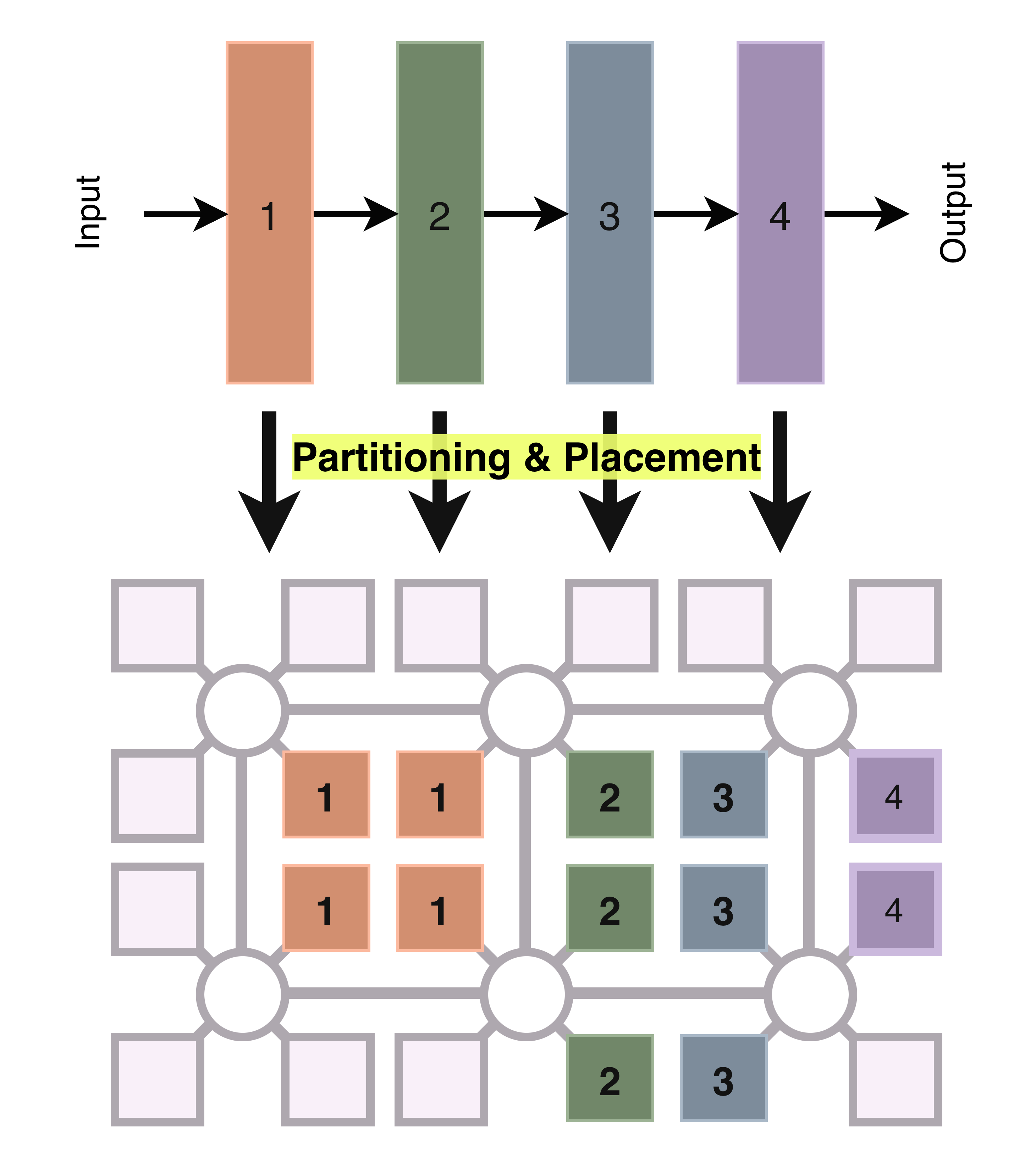}
    \caption{Partitioning and placement of a neural network to a 2D mesh of cores, interconnected by a network-on-chip. Layer 1 (orange) is partitioned to four different cores (squares), connected via routers (circles). The activations from each layer are forwarded through the routers and links.}
    \Description{Placeholder.}
    \label{fig:introduction}
\end{figure}

Spatial, or \textit{dataflow}, accelerators have emerged as promising platforms for accelerating neural network inference due to their event-driven computation, distributed memory, and tightly coupled communication fabric \cite{reuther-hpec20a}.
This hardware landscape broadly spans several architectures, at different stages of deployment, such as Google Tensor Processing Units \cite{jouppi-can17a}, Cerebras Wafer-Scale Engine \cite{lie_inside_2024}, Groq Tensor Streaming Processor \cite{abts_think_2020}, GraphCore Intelligence Processing Unit \cite{jia-arxiv19a}, as well as neuro-inspired platforms such as IBM TrueNorth \cite{akopyan_truenorth_2015}, and Intel Loihi 2 \cite{davies_advancing_2021}.

These systems offer from hundreds to millions of interconnected cores arranged in a spatial mesh that supports low-latency activation routing and energy-efficient updates.
Yet, as neural network workloads grow in scale and architectural diversity, achieving high performance on spatial accelerators increasingly depends on \emph{how} the network is \emph{partitioned} and \emph{placed} across the physical substrate, not merely \emph{which} model is being deployed \cite{james-ispd20a}. An example of a partitioning and placement is shown in Figure~\ref{fig:introduction}, where a neural network with four layers is decomposed into multiple parts and mapped onto individual cores.

In such a setting, each layer can be divided in many different ways, and each resulting partitioning can be assigned to many possible physical locations. The combination of these choices gives rise to a rapidly growing number of feasible mappings, even for moderately sized workloads. As a result, the problem is inherently combinatorial.
Each neuron and synaptic connection imposes memory and compute requirements that must fit within the individual core budgets, while the spatial location of assigned cores determines routing distances, traffic, and ultimately affects the system's throughput and energy consumption.

Current deployment pipelines \cite{yik-arxiv25a} rely heavily on manually crafted heuristics, expert knowledge of hardware topology, or analytically derived performance models.
While these approaches are valuable, they often struggle with three fundamental issues. 
First, workloads can dynamically vary, e.g., activation sparsity, and the subsequent multiply-and-accumulate load, can change over time.
Second, architectures can be heterogeneous, so that differences across revisions or multi-chip form factors complicate generalization of performance models.
Third, heuristic methods typically consider only a small subset of possible placements so that the exploration of the full design space remains limited.

These challenges motivate a black-box, hardware-in-the-loop optimization framework that can automatically search over valid partitions and placements while being guided by actual evaluation of performance metrics on the hardware in question.
Evolutionary computation provides a natural fit: it is well-suited to large, discrete, non-convex design spaces and requires no tractable gradients or analytic models of hardware behaviour.
Furthermore, evolutionary methods represent the state of the art when it comes to multi-objective optimization, allowing to search for Pareto-optimal solutions in the case of multiple metrics of interest, such as latency and energy consumption.

\paragraph{Contributions}
In this work, we propose an automated framework for mapping neural networks onto spatial accelerators, with Intel Loihi 2 as a representative testbed for conducting experiments.
Our primary contributions are as follows: \begin{enumerate}
    \item We formulate spatial accelerator deployment as a bilevel evolutionary optimization problem that jointly optimizes partitioning and placement.
    \item We introduce the first evolutionary hardware-in-the-loop mapping framework for spatial accelerators to directly minimize inference latency.
    \item We empirically validate the approach by optimizing the deployment of two workloads on a representative spatial accelerator and demonstrate substantial latency and energy improvements over existing state-of-the-art heuristics across single- and multi-chip systems.
\end{enumerate}

%% file: Chapters/2_Background.tex
\section{Background}

Spatial accelerators -- such as Intel's Loihi family of neuromorphic systems -- execute neural networks on distributed cores that are tightly coupled with on-chip routing fabrics. The performance of such systems is jointly determined by computing, memory, and communication behavior, making deployment sensitive to how a network is partitioned and spatially distributed across cores.

\subsection{Optimization for Hardware Performance}

Performance tuning for specialized accelerators is typically formulated as a design space exploration problem, in which parameters for computing power, memory, and data movement interact in non-trivial ways. On \emph{dense} deep neural network (DNN) accelerators, frameworks such as auto-tuners and compiler backends integrate analytical cost models with search algorithms to optimize tiling, scheduling, and data flow \cite{james-ispd20a,li-aspdac21a}. These methods show that hardware-efficient execution often arises from non-obvious configurations that might be overlooked by humans or static heuristics.

The mapping of deep neural networks onto spatial accelerators has been extensively investigated using analytical models and automated design space exploration \cite{zhu-arxiv25a}. Frameworks such as Timeloop~\cite{parashar-ispass19a} and subsequent accelerator-specific systems~\cite{kwon-icassplos18a,geng-micro19a,xiao-isca21a,jin-intcomp22a} explore data flow, tiling, or scheduling options using models and heuristic search. More recent approaches integrate gradient-based or GA-based optimization for mapping and hardware co-design, as in Mind Mappings~\cite{hedge-icassplos21a}, GAMMA~\cite{kao-iccad20a}, DiGAMMA~\cite{kao-date22a}, and related frameworks \cite{cai-hpca23a,xu-arxiv24b}. Unlike our work, competitors require dense DNN accelerators and analytical performance estimates.

On neuromorphic systems, which are meant for the deployment of sparse DNNs, performance depends not only on arithmetic throughput but also on spike traffic patterns, routing distances, and core-level memory capacities. Recent runtime and floorline models show that workloads can operate in compute-, memory-, or traffic-bound regimes \cite{yik-arxiv25a}. Furthermore, spatial placement decisions strongly influence end-to-end latency and energy draw~\cite{yik-arxiv25a,timcheck-arxiv26a}. This motivates treating mapping as a hardware-aware optimization problem rather than a functional compilation step, since without strong theoretical assumptions on the behavior of the neural network, anticipating the corresponding regime is difficult.


\subsection{Search Over Combinatorial Spaces}

The space of possible hardware mappings is combinatorial: for a given workload, one must \begin{enumerate*}
    \item decide how neurons or layers are partitioned into core-sized units and
    \item how these units are placed on a physical 2D mesh.
\end{enumerate*}
Both decisions are subject to discrete feasibility constraints, including per-core memory limits, global core budgets, and topology-dependent communication costs. Moreover, their combined effect on latency and energy consumption becomes apparent only after compilation and execution, resulting in objective functions that are non-convex, non-smooth, and expensive to evaluate since every evaluation is carried out on the hardware.

These characteristics render gradient-based or model-driven optimization difficult to apply reliably. Small changes in a mapping can yield disproportionate variations in routing congestion or synchronization behavior, leading to highly irregular performance landscapes. In addition, the objective function is typically piecewise constant over large regions of the search space, punctuated by sharp discontinuities when feasibility constraints are violated.

Evolutionary algorithms  are well suited to this setting because they operate directly on discrete representations, naturally accommodate hard constraints, and support structured variation operators~\cite{cicirello-arxiv23a,kao-iccad20a,kao-date22a}. In particular, population-based search enables the parallel exploration of diverse partitioning-placement trade-offs, while elitism and recombination facilitate the preservation and progressive refinement of high-quality substructures. This is especially relevant in our setting, where favorable spatial locality patterns can often be transferred across related partitionings.

While other methods exist for combinatorial optimization, in this work, we adopt a nested evolutionary strategy that explicitly mirrors the hierarchical structure of the mapping problem. Partitioning and placement are optimized by separate but coupled populations, and reordering operators are used to transfer locality information across generations, enabling efficient reuse of previously discovered spatial (sub-)structures.

\subsection{The Intel Loihi 2 Architecture}
\label{sec:loihi}

Intel Loihi~2 is a second-generation neuromorphic system consisting of neuromorphic cores arranged in a 2D mesh, with each core tightly integrating neural state, synaptic memory, and spike routing capabilities. Cores communicate via on-chip routers that support address-event routing for spikes, with communication distance affecting latency and energy. Systems scale from single-chip form factors to multi-chip boards connected via off-chip links.

From a workload deployment perspective, the architecture imposes three key constraints: \begin{enumerate*}
    \item per-core limits on neuron counts and synaptic memory,
    \item a fixed spatial topology that determines communication distances, and
    \item a global core budget that determines possible partitions.
\end{enumerate*}
As a result, mapping spiking neural networks (SNNs) to Loihi~2 requires assigning neurons to cores under capacity constraints and spatially arranging these cores to reduce routing-related congestion and latency.

This architectural context naturally leads to a mapping problem with both discrete feasibility constraints and hardware-driven performance targets, which motivates the optimization approach chosen in this work.

Previous work on neuromorphic systems has focused more on performance modeling than on automated mapping. Runtime models for Loihi 2~\cite{timcheck-arxiv26a} and floorline analyses~\cite{yik-arxiv25a} characterize computational and data traffic bottlenecks and emphasize the importance of spatial placement. Application studies demonstrate Loihi~2's capabilities for optimization and sequence modeling~\cite{smith-iclr23a,pierro-arxiv24a,pierro-icml25a} but without automatically optimizing the mapping.

In contrast, we treat partitioning and placement on Loihi~2 as a black-box mapping problem and perform hardware-guided evolutionary search. To our knowledge, no previous work combines \begin{enumerate*}
    \item[(i)] explicit combinatorial representations for partitioning and placement,
    \item[(ii)] hardware-in-the-loop fitness evaluation, and
    \item[(iii)] neuromorph-specific feasibility constraints.
\end{enumerate*}

%% file: Chapters/3_ProblemDefinition.tex
\section{Problem Definition}
\label{sec:definition}

In the following section, we formalize the partitioning and placement problem for the Intel Loihi 2 architecture, and lay the ground for the search space representation and optimization strategy introduced in \Cref{sec:algorithm}.
We then provide intuition into how mapping decisions affect latency and energy consumption.

\subsection{Partitioning and Placement}

A \textbf{workload} can be described as a tuple \( (n_1, \dots, n_L) \), where $L$ is the number of layers and \(n_i\) is the number of neurons in the \(i\)-th layer.
Each neuron has specific resource requirements, such as internal buffers and the weights of incoming connections, depending on the layer type (e.g., RNN cell or ReLU activation) and the overall workload's topology.
As no sparsity structure is imposed on the workload by default, the number of non-zero incoming connections and activation patterns may vary significantly within a layer.

As part of the mapping process, a \textbf{partitioning} is defined by splitting the neurons in each layer into one or multiple subsets, where each subset is mapped to a single virtual core.
To be admissible, a partitioning must satisfy core-level constraints, specifically the maximum limit of \num{8192} neurons per core and the memory capacity required for neuron states, spike buffers, and synaptic weights.
Thus, for a given workload, a \textbf{minimum partitioning} is denoted as \( (c_1^\text{min}, \dots, c_L^\text{min}) \), where each layer is assigned to the minimum number of cores required to satisfy these constraints.
Following previous best practice \cite{yik-arxiv25a}, we introduce two simplifying assumptions on the structure of the partitioning.
In particular, for each layer:
\begin{enumerate*}
    \item all neurons are uniformly distributed across the assigned cores, and
    \item the distribution follows the natural ordering of the neuron dimensions.
\end{enumerate*}
Although these assumptions are not strictly required by the Loihi~2 architecture or its compilation toolchain, we leave the investigation of the more general case for future work.

Let $C_\text{min}$ denote the total number of cores required by the minimum partitioning, and let $C_\text{tot}$ be the total number of available cores. Then, the number of possible admissible partitionings is
\begin{equation*}
    N_\text{partitioning} = \binom{C_\text{tot}-C_\text{min}+L}{L} = \frac{(C_\text{tot}-C_\text{min}+L)!}{(C_\text{tot}-C_\text{min})!L!},
\end{equation*}
which, even for a small \num{10}-layer workload on a single chip with \num{152} total cores exceeds $10^{13}$ possible partitionings.

Given an admissible partitioning, a \textbf{placement} assigns each required core a physical location in the system.
In the Loihi~2 mesh, described in \Cref{sec:loihi}, we define a \textbf{core location} as a triple $(x,y,c)$, where $x$ and $y$ identify the router, and $c\in\{1, \dots, 4\}$ identifies the specific core.
In multi-chip systems, an additional index is introduced to identify the chip.
The number of possible placements for a partitioning requiring $C_\text{used}$ cores is:
\begin{equation*}
    N_\text{placement}(C_\text{used}) = \frac{C_\text{tot}!}{(C_\text{tot}-C_\text{used})!},
\end{equation*}
as the order of the unused cores has no effect on performance. Assigning all cores, yields the maximum number of possible placements exceeding $10^{267}$, whereas a workload occupying only half of the cores still induces a search space of size $>10^{155}$. 


\subsection{Effects on Latency and Power}

The configuration of partitioning and placement has a major impact on latency and power consumption.
On Loihi~2, optimizing partitioning improves latency up to $1.73\times$ for convolutional workloads~\cite{yik-arxiv25a}, while placement alone can induce up to a $6\times$ performance gap for linear layers~\cite{timcheck-arxiv26a}.
These results highlight the importance of optimizing the mapping as a whole.

Partitioning and placement often interact in complex and counteracting ways. Increasing parallelism by distributing a compute-bottlenecked layer across more cores can reduce per-core workload and amortize static power, but it also raises inter-core communication, which may saturate the routing fabric and increase dynamic power consumption.
Meanwhile, traffic bottlenecks can potentially be mitigated by better placements. Depending on the location of the traffic bottleneck and the location of otherwise underutilized bandwidth on the routing fabric, improvements can be obtained by adapting the placement. For instance, a linear layer placed in an X-shaped pattern can reduce the traffic bottleneck up to $6x$\cite{timcheck-arxiv26a}. 
Alternatively, more spatially localized patterns can help disentangle otherwise overlapping communications from different layers and reduce overall communication routing distances.
Considering such patterns can help reduce and balance routing fabric load and reduce average routing distances, thereby alleviating congestion and reducing execution time and/or power.

At the same time, feasible placements are constrained by the finite size and topology of the mesh, and favorable patterns may no longer be realizable once the workload exceeds certain spatial limits. 
As a result, improving partitioning often shifts the dominant bottleneck to communication, while improving placement may expose new compute or capacity constraints. This iterative trade-off between parallelism, communication, and spatial locality is characteristic of neuromorphic mappings and becomes even more pronounced for deeper networks and multi-chip deployments. Effective optimization thus requires joint handling of partitioning and placement rather than treating them as independent stages.


%% file: Chapters/4_Algorithm.tex
\section{Evolving Neural Network Mappings}
\label{sec:algorithm}

In this section, we introduce an evolutionary algorithm for the optimization of partitioning and placement. To this end, we propose genetic representations for partitionings and placements in \Cref{sec:genetic-representation}, describe mutation and reordering operators in \Cref{sec:approach-operators}, and devise a nested evolution algorithm approach in \Cref{sec:approach-nea}.

\subsection{Genetic Representation}
\label{sec:genetic-representation}

Following the notation in \Cref{sec:definition}, we adopt a decoupled representation in which partitioning and placement are encoded separately.

\paragraph{Partitioning genotype}
We first define a \textbf{partitioning genotype} that specifies how the available cores are distributed across layers beyond the minimum feasible allocation. Let $C_\text{min}$ denote the number of cores required by the minimum partitioning. Any additional cores can be flexibly assigned to layers to increase parallelism.

We encode this allocation as a vector:
\begin{align*}
    \mathbf{x} &= (x_1,\dots,x_L,C_\text{unused}) \in\mathbb{N}_0^{L+1}, \\
    \text{where } C_\text{unused} &= C_\text{tot}-C_\text{min}-\sum_{i=1}^L x_i \notag
\end{align*}
Here, each $x_i$ represents the number of additional cores assigned to layer $i$, while the final element $C_\text{unused}$ tracks the number of remaining unallocated cores. Explicitly encoding $C_\text{unused}$ simplifies feasibility checking during evolution, as all individuals automatically satisfy the global core budget.

\paragraph{Placement genotype}
We represent a \textbf{placement genotype} as a permutation of all physical cores available in the system, independently of whether they are used by the current mapping:
\begin{equation}
    \boldsymbol\omega = (\omega_1,\dots,\omega_{C_\text{tot}}),
\end{equation}
where each element $\omega_i=(x_i,y_i,c_i)$ denotes a specific core location. 

\paragraph{Construction of a complete mapping}
A full mapping is obtained by combining a partitioning genotype $\mathbf{x}$ with a placement genotype $\boldsymbol\omega$. For each layer $i$, we assign $c_i^{\min} + x_i$ consecutive cores from the permutation to that layer, where $c_i^{\min}$ denotes the minimum number of cores required by layer $i$. $C_\text{unused}$ cores remain unassigned.

\subsection{Mutation and Reordering Operators}
\label{sec:approach-operators}

\begin{figure}[t]
    \centering
    \includegraphics[width=\linewidth]{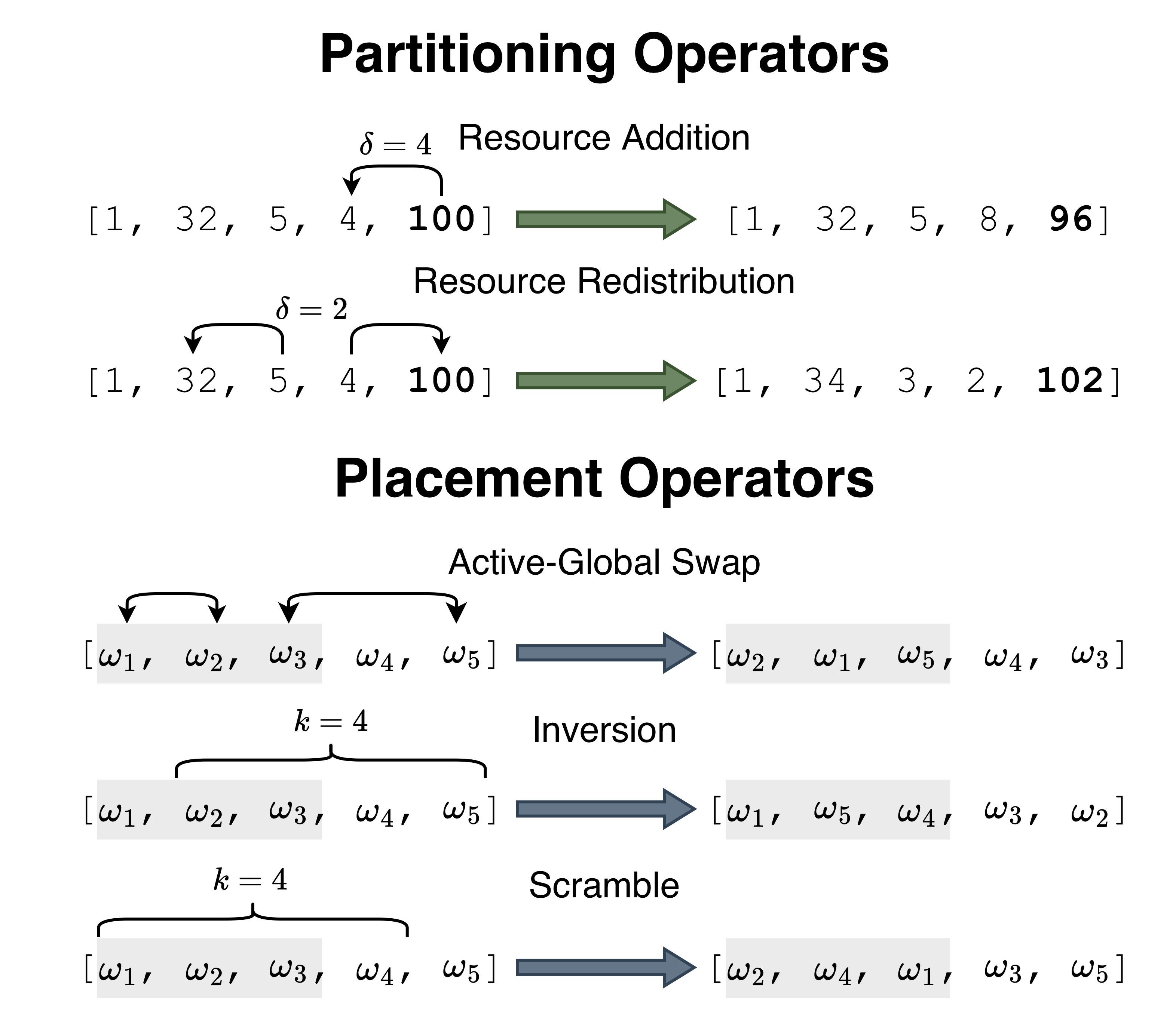}
    \caption{Illustration of mutation and reordering operators for the partitioning and placement populations respectively.}
    \Description{Placeholder.}
    \label{fig:mutation-operators}
\end{figure}

We adopt mutation operators that preserve individual feasibility and introduce a reordering operator to facilitate fitness transfer across different partitionings.
A visualization of the operators is provided in \Cref{fig:mutation-operators} as an overview.

For the partitioning population, an offspring is generated via the following mutation procedure.
Each gene, except for the last one tracking parity, is independently selected for mutation with probability $p_\text{part}^\text{mut}$, either by:
\begin{description}
    \item[Resource addition:] With probability $p_\text{part}^\text{add}$, a random number $\delta$ of additional cores is added to the layer from the pool of unused cores.
    \item[Resource redistribution:] With probability $p_\text{part}^\text{swap}=1-p_\text{part}^\text{add}$, a random number $\delta$ of cores is moved from the current layer to another layer or to the unused pool.
\end{description}
In both cases, the amount of moved cores $\delta$ is uniformly sampled from $\{1,\dots,\delta_\text{max}\}$ and capped to the number of available cores in the unused group to maintain feasibility.
The ordering bias of the operators is averaged out by randomizing the order of update.

For the placement population, we rely on previous research on permutation operators \cite{cicirello-arxiv23a} to design a mutation strategy.
Since placement individuals are always evaluated in the context of a specific partitioning, we account for the total number of used cores $C_\text{used}$, which correspond to the first $\omega_1,\dots,\omega_{C_\text{used}}$ elements in $\boldsymbol\omega$.
An offspring is generated by applying one of the following mutations:
\begin{description}
    \item[Active-global swap:] With probability $p^\text{swap}_\text{place}$, perform $\frac{k}{2}$ swaps between pairs $\omega_i$ and $\omega_j$, with $i$ and $j$ being uniformly drawn from $\{1,\dots,C_\text{used}\}$ and $\{1,\dots,C_\text{tot}\} \setminus \{i\}$ respectively.
    \item[Inversion:] With probability $p^\text{inverse}_\text{place}$, reverse a substring of length $k$ starting at a random index $i\in\{1,\dots,C_\text{used}\}$.
    \item[Scramble:] With probability $p^\text{}_\text{place}=1-p^\text{scramble}_\text{place}-p^\text{inverse}_\text{place}$, randomly shuffle a sequence of length $k$ starting from a random index $i\in\{1,\dots,C_\text{used}\}$.
\end{description}
For all three cases, the magnitude of the mutation $k$ is uniformly drawn from $\{1,\dots, \lfloor \alpha C_\text{used} \rfloor \}$, with the parameter $\alpha$ specified at initialization, allowing it to scale with the number of active cores.

Finally, we introduce a \textbf{reordering operator} with the objective of facilitating the fitness transfer of a placement $\boldsymbol\omega$ optimized for partitioning $\mathbf{x}$ to a different partitioning $\mathbf{x}'$.
The operator produces a placement $\boldsymbol\omega'$ which minimizes the mismatch between the physical cores assigned to each layer across the two configurations. Specifically, the operator constructs $\boldsymbol\omega'$ by iterating through each layer~$i$ and preserving the subset of physical cores originally assigned to it in $\boldsymbol\omega$.
If the new partitioning $\mathbf{x}'$ allocates fewer cores to layer~$i$ than $\mathbf{x}$ (i.e., $x'_i < x_i$), the first $x'_i$ cores from the original assignment are retained, and the surplus is released to the unallocated pool.
Conversely, if the allocation increases (i.e., $x'_i > x_i$), the original assignment is preserved in its entirety, and the deficit is filled by drawing from the unallocated pool.
This strategy ensures that the spatial locality learned for specific neural network layers is retained, allowing the placement optimization to adapt to partitioning changes without disrupting the geometric structure of the placement obtained in a preceding generation.

\begin{figure}
    \centering
    \includegraphics[width=\linewidth]{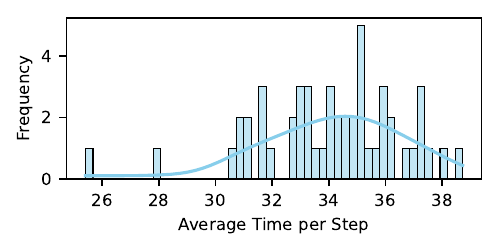}
    \caption{Distribution of average latency for a fixed partitioning across 50 random placements. Results are on a 6-layer MLP workload, averaged over 200 time steps.}
    \Description{Placeholder.}
    \label{fig:fitness-var}
\end{figure}

\subsection{Nested Evolution Algorithm}
\label{sec:approach-nea}

We focus on minimizing end-to-end latency, measured as the \textbf{average time per step} in microseconds (\unit{\us}). We denote this metric by $T_\text{step}(\mathbf{x}, \boldsymbol\omega)$ to emphasize its dependence on both partitioning $\mathbf{x}$ and placement $\boldsymbol\omega$.
The fitness function that is sought to be maximized is therefore defined as $-T_\text{step}(\mathbf{x}, \boldsymbol\omega)$. 

A natural approach would be to optimize partitioning and placement independently using separate populations. However, as shown in \Cref{fig:fitness-var}, the performance of a given partitioning exhibits substantial variance across different placements. Consequently, reliable evaluation of partitioning quality requires multiple placement-specific measurements, making naive decoupling ineffective.

Motivated by iterative tuning strategies adopted by experts~\cite{yik-arxiv25a}, we formulate the mapping problem as a bilevel optimization:
\begin{equation}
    \max_{\mathbf{x}} -T\left(\mathbf{x}, \operatorname*{arg\,max}_{\boldsymbol\omega} -T(\mathbf{x},\boldsymbol\omega)\right),
\end{equation}
where the lower level optimizes the placement for a fixed partitioning, and the upper level optimizes the partitioning itself.

We realize this formulation using a nested evolutionary strategy (ES), employing a $(1+\lambda)$-ES at both levels.
Each generation starts from a current solution $(\textbf{x}_t, \boldsymbol\omega_t)$ and proceeds in two stages.
During the \textbf{partitioning evolution step}, a number of partitioning offsprings $\lambda_\text{part}$ are generated through mutation from $\mathbf{x}_t$.
Each offspring $\mathbf{x}_i'$ is evaluated as $(\mathbf{x}_i', \boldsymbol\omega_i')$, where $\boldsymbol\omega_i'$ is obtained by aligning $\boldsymbol\omega_t$ from $\mathbf{x}_t$ to $\mathbf{x}_i'$ with the reordering operator introduced in the previous section.
An elitist selection is then applied to obtain the new partitioning $\mathbf{x}_{t+1}$, while the associated placement, being it $\boldsymbol\omega_t$ or $\boldsymbol\omega_i'$, is maintained as $\bar{\boldsymbol\omega}_t$.
The \textbf{placement evolution step} then proceeds with the generation of $\lambda_\text{place}$ placement offsprings $\boldsymbol\omega_i'$ through mutation of $\bar{\boldsymbol\omega}_t$, which are evaluated as $(\mathbf{x}_{t+1},\boldsymbol\omega_i')$.
Elitist selection is then applied to obtain the new individual $(\mathbf{x}_{t+1},\boldsymbol\omega_{t+1})$.



\subsubsection{Initialization Heuristics}
\label{sec:initialization}

\begin{figure}
    \centering
    \begin{subfigure}[t]{0.47\linewidth}
        \centering
        \fbox{\includegraphics[width=.95\linewidth]{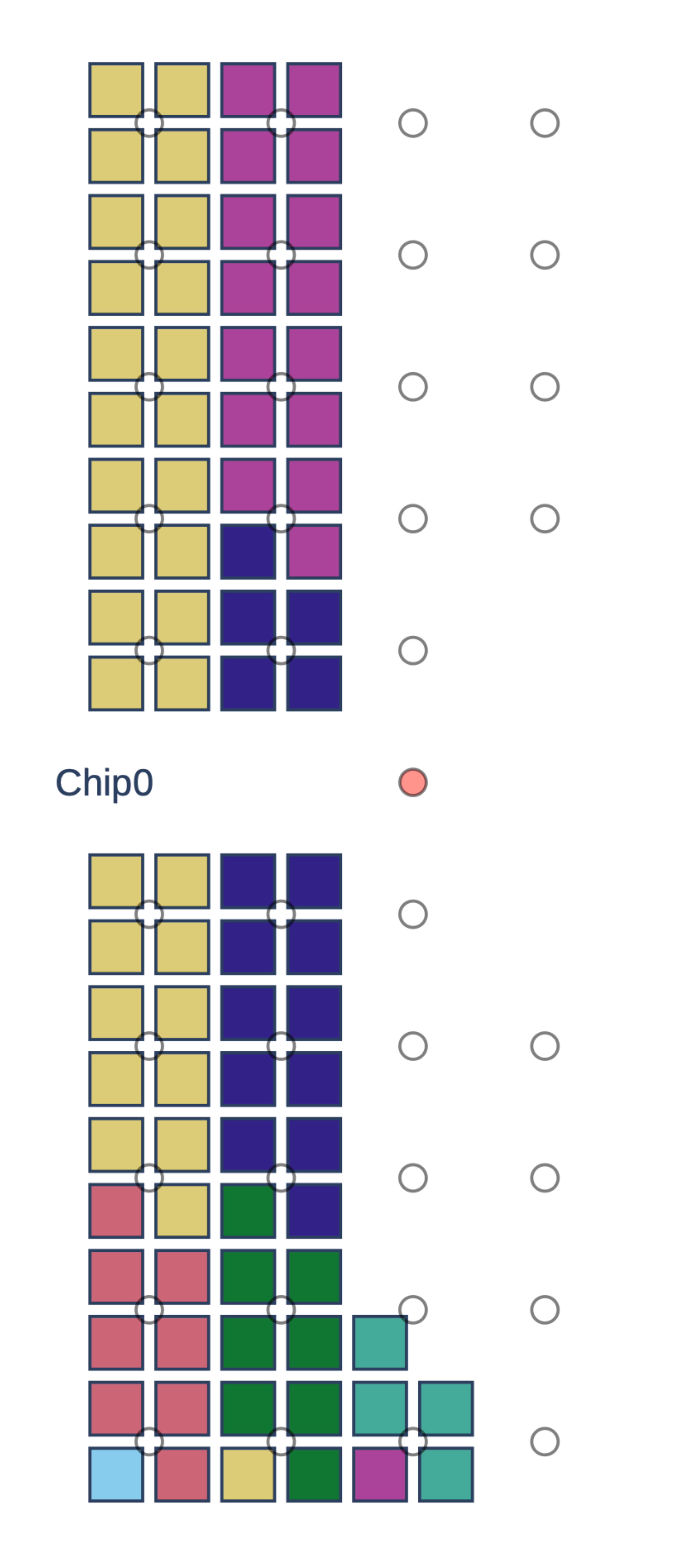}}
        \caption{Packed Column-Wise}
    \end{subfigure}
    \hfill
    \begin{subfigure}[t]{0.47\linewidth}
        \centering
        \fbox{\includegraphics[width=.95\linewidth]{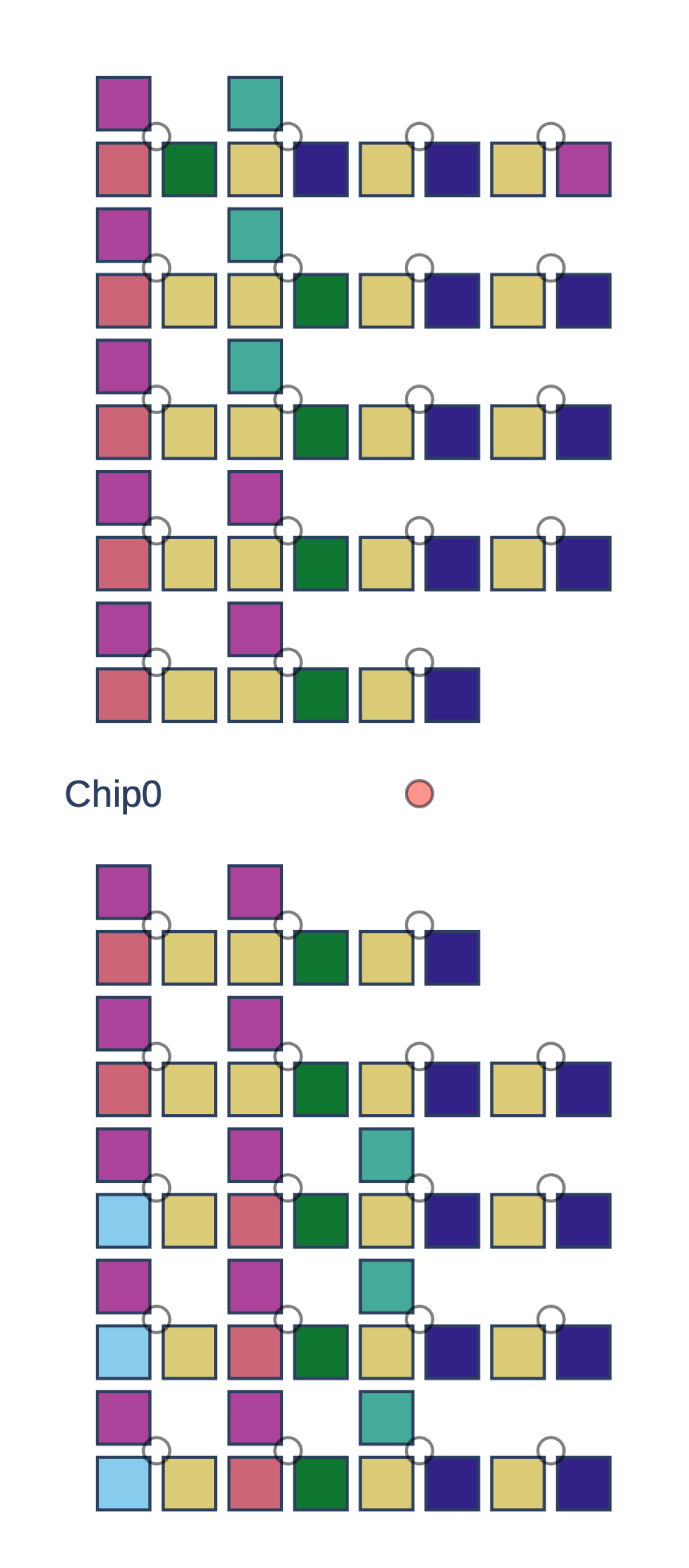}}
        \caption{Spread Column-Wise}
    \end{subfigure}
    \caption{Diagram of placement heuristics on SparseMLP\text{-}1.}
    \Description{Placeholder.}
    \label{fig:initialization}
\end{figure}

We define a few structured mapping configurations that we use to initialize the evolution and compare the results obtained with our approach.
Starting from the minimum partitioning, we define a \textit{min + k} partitioning, where $k$ additional cores are uniformly assigned to each layer.
For placement, we evaluate four different heuristics, combining two filling orders, column-major and row-major, and two different filling granularities, packed and spread.
For clarity, we report packed column-major and spread column-major configurations in \Cref{fig:initialization}.
The spread configuration starts by utilizing only one core for each router, going through all the routers before adding a second one.
The packed configuration, on the other side, saturates all the four cores of each router before moving to the next one.
These two different configurations lead to significantly distinct traffic patterns, which can be more or less beneficial based on the specifics of the workload \cite{timcheck-arxiv26a}.

%% file: Chapters/5_Results.tex
\section{Experimental Results}

\begin{figure*}[h!]
    \centering
    \begin{subfigure}[t]{0.48\textwidth}
        \centering
        \includegraphics{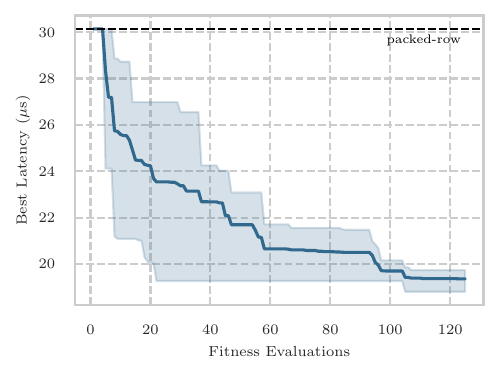}
        \caption{SparseMLP-1}
        \label{fig:sparsemlp-1}
    \end{subfigure}
    ~ 
    \begin{subfigure}[t]{0.48\textwidth}
        \centering
        \includegraphics{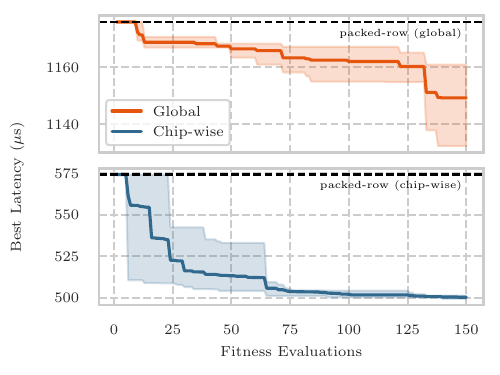}
        \caption{SparseMLP-2}
        \label{fig:sparsemlp-2}
    \end{subfigure}
    \caption{Progression of the best latency found by the nested evolution algorithm over the number of fitness evaluations for \textbf{(a)} SparseMLP\text{-}1 workload on a single chip, and \textbf{(b)} SparseMLP\text{-}1 workload on two chips. The results are averaged over 5 random trials for each workload, with the shaded area covering the min/max interval.}
    \Description{Placeholder.}
\end{figure*}

In this section, we evaluate our proposed evolutionary strategy for the bi-level optimization problem. First, we describe the setup and the workloads benchmarked in \Cref{sec:exp-setup}. Results for single-chip and multi-chip mappings are presented in \Cref{sec:exp-single,sec:exp-multi} respectively. We further analyze the latency-power landscape in \Cref{sec:exp-lat-power} and ablate design components of our approach (\Cref{sec:exp-ablation}). In \Cref{sec:exp-mappings}, we analyze the phenotype of evolved mappings .

\subsection{Setup and Workloads}\label{sec:exp-setup}

We consider two sparse multi-layer perceptron (MLP) workloads:
\begin{description}
    \item[SparseMLP-1:] A 6-layer MLP with hidden dimensions ranging from 512 to 4096 neurons. The model consists of a total of \qty{16.8}{M} parameters.
    \item[SparseMLP-2:] A 12-layer MLP with hidden dimensions ranging from 1024 to 4096 neurons. The model consists of a total of \qty{33.6}{M} parameters in total.
\end{description}

Both models have input size \num{1024} and \num{512} as output size exhibit $85\%$ unstructured weight sparsity, while activation sparsity increases linearly from $15\%$ in the first layer to $80\%$ in the last layer.

Experiments were conducted on a Kapoho Point system with eight N3D1-revision Loihi~2 chips running NxCore 2.5.18 and NxKernel 0.4.0. Latency is measured as average time per inference step over 200 iterations, and power is measured using on-board sensors.




\subsection{Results on Single-Chip Mapping}\label{sec:exp-single}

We evaluate our nested ES on mapping the SparseMLP\text{-}1 to a single Loihi 2 chip.
The results are reported in \Cref{fig:sparsemlp-1}.
We initialize the search with the \textit{min + 2} partitioning and evaluate the four placement heuristics, described in \Cref{sec:initialization}, plus a random placement.
The best configuration is used as the initial parent for evolution, and a total of \num{125} fitness evaluations are performed.

We observe that the packed row-major configuration provides the best performance among heuristics with a \qty{30.14}{\mu\s} latency, a significant gap from spread column-major (\qty{40.76}{\mu\s}), packed column-major (\qty{34.86}{\mu\s}), and spread row-major (\qty{32.58}{\mu\s}).
Starting from there, our nested EA improves performance and consistently converges at an average best latency of \qty{19.36}{\mu\s}, improving by $35.8\%$ wrt.~packed row-major.
Evolution still finds improved solutions up to $85\%$ of the evaluation budget, hinting at further improvements with longer runtimes.

From the user perspective, each fitness evaluation requires approximately \qty{5}{\s}, which results in a total execution time for evolution of \qty{10}{\minute}.
We consider this practical for real-world usage, given the optimization has to be performed only once per workload for subsequent long-time deployments.

\subsection{Results on Multi-Chip Mapping}\label{sec:exp-multi}

We investigate two generalizations of our approach for deploying SparseMLP-2 across multiple chips: a global strategy and a chip-wise strategy.
In the global strategy, we do not distinguish between the different chips, but rather combine the available cores and let the evolutionary strategy choose from all available cores. 

The chip-wise strategy, on the other hand, first divides the workload into segments, which are then distributed across the individual chips and the mapping is then optimized separately for each chip. More specifically, instead of a single search, we therefore execute the proposed evolutionary strategy for each segment individually.

Figure~\ref{fig:sparsemlp-2} shows that the global strategy is significantly inferior to the chip-wise strategy, exhibiting substantially higher latency. This clearly indicates that the larger search space is more difficult to manage and that a divide-and-conquer principle is advantageous here. While the resulting latency is superior, it also increases the cost of deployment. Exploring strategies that can efficiently handle the increased complexity is an interesting avenue for future work.





\subsection{Analysis of Latency-Power Landscape}\label{sec:exp-lat-power}

\begin{figure}
    \centering
    \includegraphics[width=\linewidth,trim={0 0.2cm 0 0},clip]{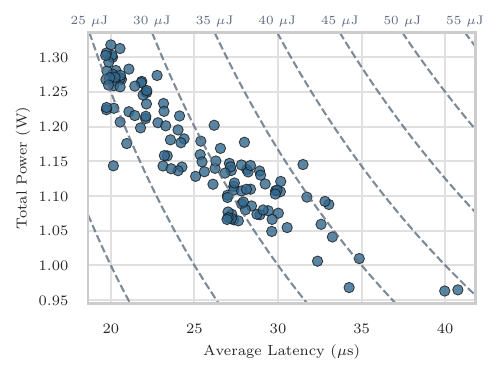}
    \caption{Evaluation of power and energy consumption for all individuals evaluated by evolution on MLP-Small. Minimizing latency increases hardware usage and power draw, but improves overall energy consumption, with the fastest configurations being also the most energy-efficient.}
    \Description{Placeholder.}
    \label{fig:latency-energy-mlp}
\end{figure}

We assess the effect of our latency optimization strategy on power consumption and profile \num{125} individuals for the SparseMLP\text{-}1 workload, evaluated in one evolutionary run.
The total power is the sum of static power, averaged over \qty{5}{s} with the board in idle mode and normalized based on chip utilization, and dynamic power, averaged over \qty{1}{M} inference steps.
The results in \Cref{fig:latency-energy-mlp} yield two insights.

Firstly, minimizing latency implicitly increases power consumption by favouring the use of more hardware resources (i.e., cores).
This results in the second slowest configuration
reaching the lowest power consumption at \qty{0.96}{\W}, while the fastest configurations
reach a power consumption of up to \qty{1.3}{\W} (a $+35\%$ gap).
However, a power variability of $\sim10\%$ is visible across all levels of latency, suggesting  scope for direct optimization of power through mapping.

Secondly, we highlight that faster configurations are also more energy efficient.
In particular, while evolution improves latency from \qty{40.76}{\mu\s} to \qty{19.7}{\mu\s}, the energy per sample drops from \qty{39.32}{\mu\J} to \qty{23.04}{\mu\J}, with an improvement of $41.4\%$.
As previously reported \cite{yik-arxiv25a}, the predominance of static power makes latency optimization the principal way to improve energy efficiency of Loihi 2 systems, which is a well-known effect on SRAM-based accelerators \cite{hennessy2017computer}.

Overall, the nuanced connections between latency, power, energy efficiency, and hardware utilization suggest significant space for future work in the direction of multi-objective and constrained optimization.
This would further enable practitioners to account for specific deployment constraints during the mapping process.

\subsection{Ablation Studies}\label{sec:exp-ablation}

\begin{figure}
    \centering
    \begin{subfigure}[t]{\linewidth}
        \centering
        \includegraphics[width=\linewidth,height=0.5\linewidth]{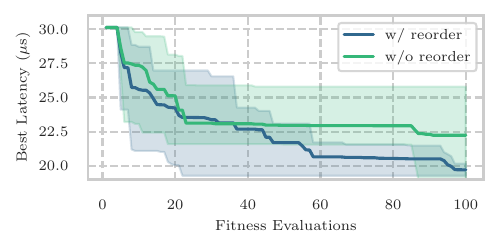}
        \caption{Effect of reordering operator}
        \label{fig:ablation-reorder}
    \end{subfigure}
    \vfill
    \begin{subfigure}[t]{\linewidth}
        \centering
        \includegraphics[width=\linewidth,height=0.5\linewidth]{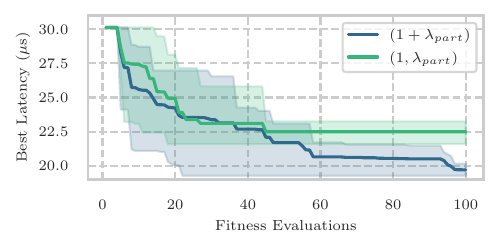}
        \caption{Effect of elitism in partitioning population}
        \label{fig:ablation-elitism}
    \end{subfigure}
    \vfill
    \begin{subfigure}[t]{\linewidth}
        \centering
        \includegraphics[width=\linewidth,height=0.5\linewidth]{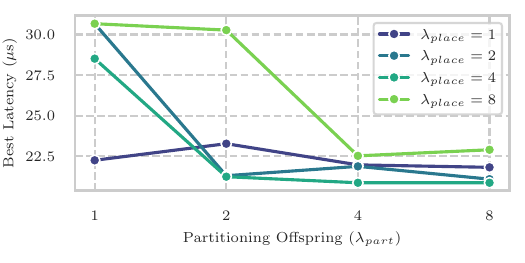}
        \caption{Effect of population sizes}
    \label{fig:ablation-lambda}
    \end{subfigure}
    \caption{Ablation study on the effect of \textbf{(a)} the reordering operator, \textbf{(b)} the use of elitism in the partitioning population, and \textbf{(c)} the population sizes. The results are averaged over 5 random trials on the SparseMLP\text{-}1 workload, with the shaded area covering the min/max interval.}
    \Description{Placeholder.}
\end{figure}


\paragraph{Effect of the reordering operator}

In \Cref{sec:algorithm}, we introduced a reordering operator to facilitate the robust fitness evaluation of partitioning offsprings.
Here, we evaluate its effect on the convergence of the algorithm, comparing the baseline evolution with a variant without the reordering operator.

\Cref{fig:ablation-reorder} shows that, without reordering, evolution settles on longer plateaus early on, resulting in a higher average best latency of \qty{22.23}{\mu\s}, a $13\%$ gap compared to including reordering reaching \qty{19.70}{\mu\s}.
In addition, comparing the evolution traces, the variant without reordering exhibits $22\%$ more partitioning generations where no improvement was found.
This suggests that the operator facilitates navigating the fitness landscape.

We also note that, within the five random trials, the variant without reordering finds the best global solution with a latency of \qty{18.68}{\mu\s}, compared to the \qty{19.26}{\mu\s} solution found by the baseline.
Hence, while the reordering operator improves stability, robust fitness evaluation at the upper level remains challenging  \cite{deb-gecco24a}.

\paragraph{Effect of elitism in partitioning}

We evaluate the influence of elitism in the partitioning population by comparing a $(1+\lambda)$ strategy with a non-elitist $(1, \lambda)$ variant, in which parent individuals are discarded regardless of the quality of their offspring.
Figure~\ref{fig:ablation-elitism} shows that removing the elite significantly impairs the convergence speed and the final solution quality. Without preserving the elites, high-quality partitions are often lost, which in turn hinders progress and leads to premature convergence.
These results confirm that preserving strong partitioning solutions is essential for the bilevel evolutionary setting, in which noisy evaluations at the lower level otherwise lead to disruption of selection at the upper level.

\subsubsection{Effect of population sizes}
\label{sec:ablation-lambdas}

We study the effect of the offspring population size parameters $\lambda_\text{part}$ and $\lambda_\text{place}$, while keeping the overall fitness evaluation budget constant, thus resulting in a varying number of generations.
This can be interpreted as a way to control the ratio of evaluations used to solve the upper level and lower level respectively.
The results, reported in \Cref{fig:ablation-lambda}, show that increasing $\lambda_\text{part}$ from $1$ to $4$ is beneficial across all values of $\lambda_\text{place}$, with mixed effects when increasing $\lambda_\text{part}$ from $4$ to $8$.
In addition, we highlight that a balance between the two parameters is beneficial, as the configuration $\lambda_\text{part}=\lambda_\text{place}$ is always the best performing one, apart from the case of $\lambda_\text{place}=8$ which performs weakly across all values of $\lambda_\text{part}$.

\subsection{Analysis of Evolved Mappings}\label{sec:exp-mappings}
\begin{table}
    \centering
    \caption{Performance model analysis of mappings in \Cref{fig:solution-diagrams}}
    \label{tab:timing_breakdown}
    \begin{tabular}{lrrr}
        \toprule
        \textbf{Metric} & \textbf{Worst} & \textbf{Best} & \textbf{\% Improv.} \\
        \midrule
        Est. SynOps Time   & \qty{9.61}{\mu\s}   & \qty{8.10}{\mu\s}   & \qty{15.7}{\percent} \\
        Est. SynMem Time   & \qty{17.78}{\mu\s}  & \qty{15.86}{\mu\s}  & \qty{10.8}{\percent} \\
        Est. DendOps Time  & \qty{4.48}{\mu\s}   & \qty{0.68}{\mu\s}   & \qty{84.8}{\percent} \\
        Est. Link Time     & \qty{333.88}{\mu\s} & \qty{259.95}{\mu\s} & \qty{22.1}{\percent} \\
        \midrule
        \textbf{Measured Latency} & \textbf{\qty{48.03}{\mu\s}} & \textbf{\qty{19.25}{\mu\s}} & \textbf{\qty{59.9}{\percent}} \\
        \bottomrule
    \end{tabular}
\end{table}

We further analyzed solutions obtained by employing the proposed evolutionary strategy but substituting the hardware-in-the-loop fitness function by the performance model devised in \cite{timcheck-arxiv26a}. Since the performance model does not account for activation sparsity, which is a dynamic property of workloads, the performance model only provides an upper bound for the runtime of each stage based on chip statistics and hardware counters.
Since all stages of a workload run mostly asynchronously, the predicted time is the maximum time among all stages.

\begin{figure}
    \centering
    \begin{subfigure}[t]{0.46\linewidth}
        \centering
        \fbox{\includegraphics[width=.95\linewidth]{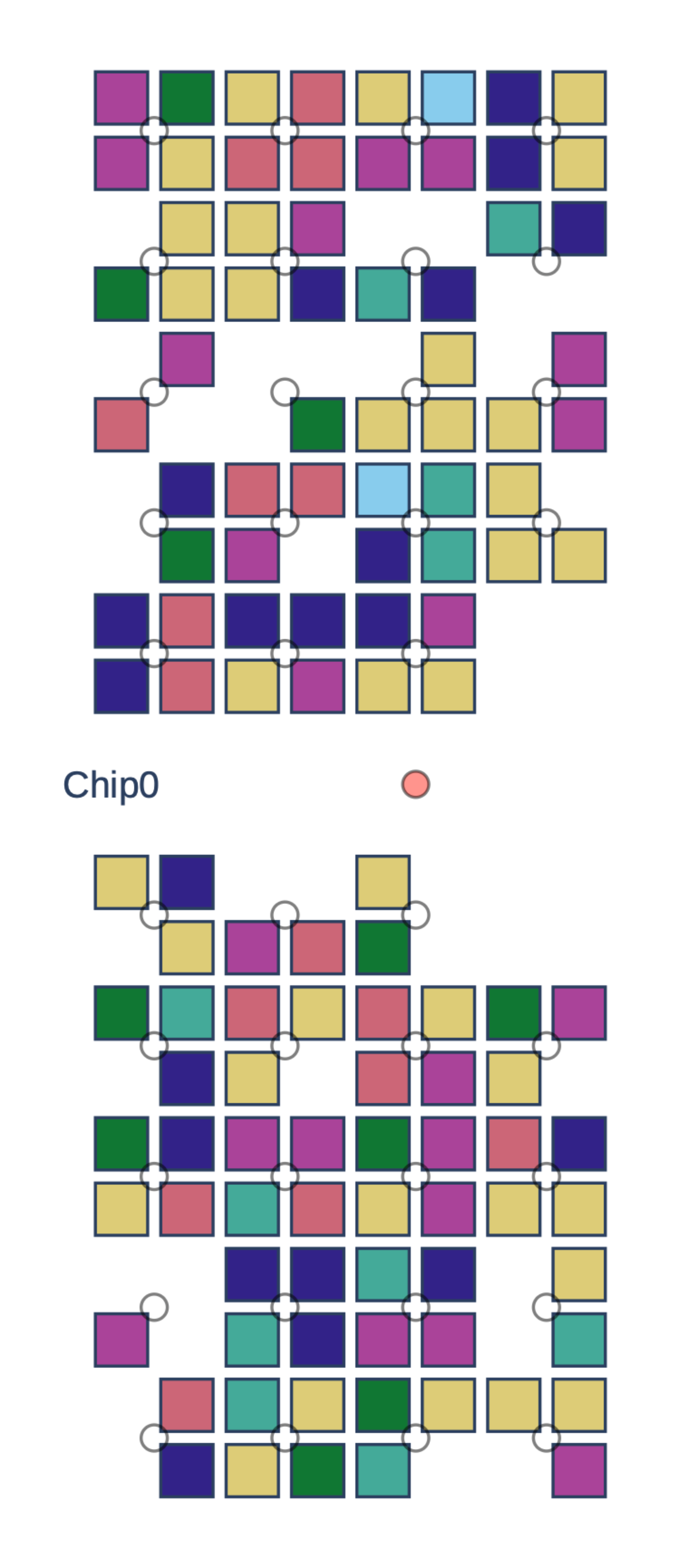}}
        \caption{Worst individual}
    \end{subfigure}
    \hfill
    \begin{subfigure}[t]{0.46\linewidth}
        \centering
        \fbox{\includegraphics[width=.95\linewidth]{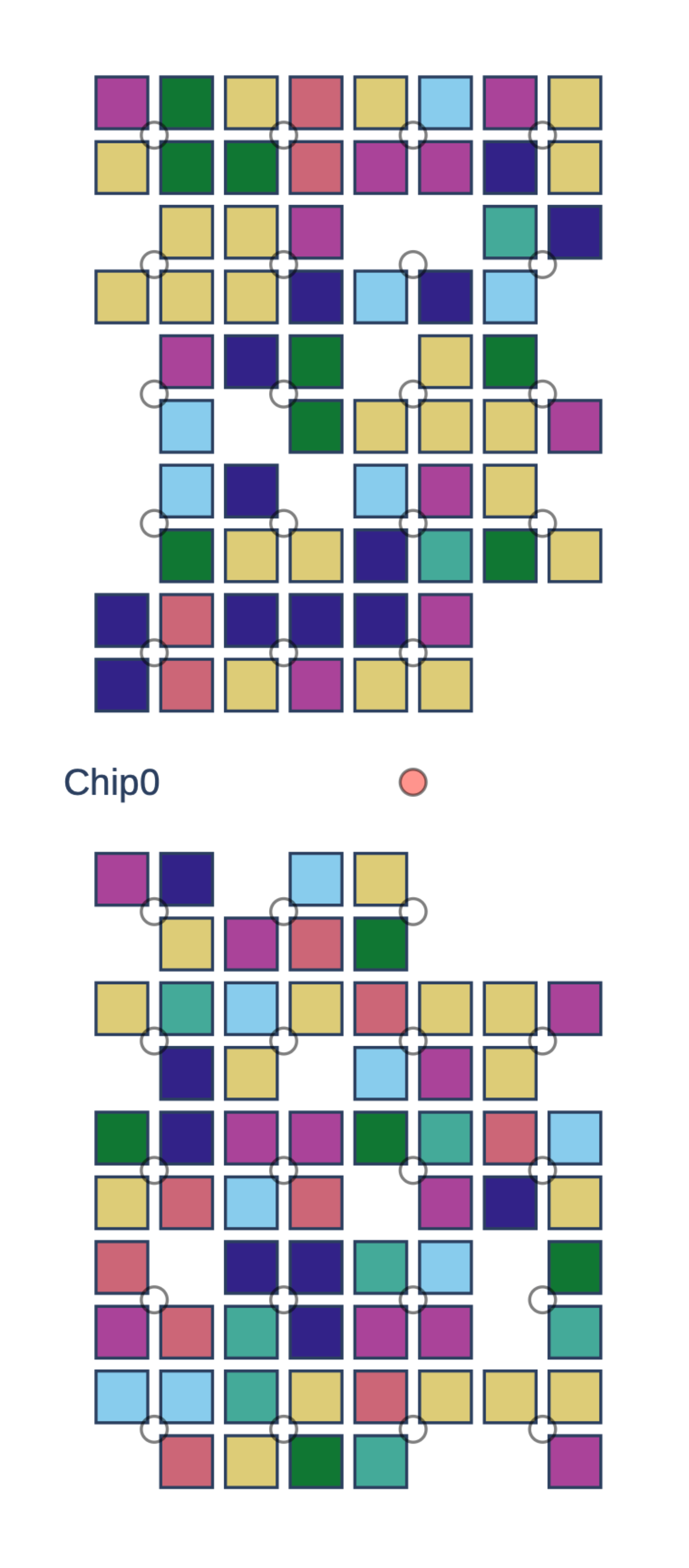}}
        \caption{Best individual}
    \end{subfigure}
    \caption{Diagrams of worst and best SparseMLP-1 mappings.}
    \Description{Placeholder.}
    \label{fig:solution-diagrams}
\end{figure}

\Cref{tab:timing_breakdown} reports the analysis for both the best and the worst SparseMLP\text{-}1 configurations, which are additionally visualized in \Cref{fig:solution-diagrams}.
Both configurations are traffic-dominated, as the \textit{link time} is the only upper bound never exceeded by the measured latency.
Although the observed latency remains well below the predicted traffic resolution time, this suggests that the performance model can only serve as a conservative fitness signal. Yet, the costs of the model are lower.
Additionally, evolutionary optimization improves runtime across all execution stages.
Given the dynamic nature of the workload, different inference steps exhibit different bottlenecks, explaining why evolution promotes improvements across all stages.

%% file: Chapters/6_Conclusions.tex
\section{Conclusion and Future Work}

We have presented an evolutionary algorithm for the joint optimization of partitioning and placement of neural workloads on spatial accelerators. By formulating the mapping as a fully discrete, hardware-aware search problem and combining tailored genotype representations with nested evolutionary search, our approach discovers mappings that significantly reduce end-to-end latency and energy consumption compared to state-of-the-art heuristic baselines. Evaluations on Loihi~2 show consistent improvements across heterogeneous workloads and highlight the importance of spatial locality, core budget allocation, and topology-aware placement for neuromorphic performance.

There are several promising avenues for future work. First, extending our methodology to larger multi-chip and multi-board implementations may require hierarchical or modular mapping strategies to ensure scalability. Second, integrating analytical runtime models or learned surrogates could reduce hardware evaluation costs and enable broader exploration of the design space. Third, the inclusion of fall-through execution modes, asynchronous workloads, and heterogeneous layer structures would extend applicability beyond the uniform pipeline execution considered here. Finally, the emergence of structured mapping behaviors offers opportunities for transfer learning and warm starts across workloads or hardware revisions.